\documentclass[a4paper]{IEEEtran}
\usepackage{graphicx} % Required for inserting images
\usepackage{authblk}
\usepackage{booktabs}
\usepackage{subcaption}
\usepackage{multirow}
\usepackage{csquotes}
\usepackage{hyperref}
\usepackage{xcolor}
\usepackage{amsmath}
\usepackage{rotating}
\usepackage{newfloat}
\usepackage{fvextra}
\usepackage{listings}

\DeclareFloatingEnvironment[
  fileext=lol,
  listname={List of Listings},
  name=Listing
]{listing}

\title{Leveraging ChatGPT's Multimodal Vision Capabilities to Rank Satellite Images by Poverty Level: Advancing Tools for Social Science Research}
\author[1]{Hamid Sarmadi}
\author[2]{Ola Hall}
\author[1]{Thorsteinn Rögnvaldsson}
\author[1,3]{Mattias Ohlsson}
\affil[1]{Center for Applied Intelligent Systems Research (CAISR), Halmstad University, Sweden}
\affil[2]{Department of Human Geography, Lund University, Sweden}
\affil[3]{Department of Earth and Environmental Sciences, Lund University, Sweden}

\begin{document}
\maketitle
\begin{abstract}
This paper investigates the novel application of Large Language Models (LLMs) with vision capabilities to analyze satellite imagery for village-level poverty prediction. Although LLMs were originally designed for natural language understanding, their adaptability to perform multimodal tasks, including geospatial analysis, has opened new frontiers in data-driven research. By leveraging advancements in vision-enabled LLMs, we assess their ability to provide interpretable, scalable, and reliable insights into human poverty from satellite images. Using a pairwise comparison approach, we demonstrate that ChatGPT can rank satellite images based on poverty levels with an accuracy comparable to a model trained on human-expert-identified image features and better than direct expert wealth classifications reported previously\cite{sarmadi_human_2024}. These findings highlight both the promise and the limitations of LLMs in socioeconomic research, providing a foundation for their integration into poverty assessment workflows. This study contributes to the ongoing exploration of unconventional data sources for welfare analysis and opens pathways for cost-effective, large-scale poverty monitoring. Our results also put forward the question of how reliable public datasets, such as DHS, are for retrieving wealth indices that are anonymized by adding noise to the location coordinates. The code and data used for the analyses in this paper are publicly available.
\end{abstract}

\flushbottom

\thispagestyle{empty}

\section{Introduction}\label{sec1}

The use of unconventional data sources to assess and predict poverty has gained significant traction in recent years, driven by the need for timely, granular insights into economic well-being. This is highly relevant for tracking progress towards the Sustainable Development Goals (SDGs). Traditional poverty assessments, such as surveys and censuses, are often costly, time-consuming, and infrequent, particularly in developing countries where resource limitations and/or political constraints hinder comprehensive data~\cite{burke_using_2021}. As a response, researchers have explored innovative data streams such as satellite imagery~\cite{hersh_open_2021, sarmadi_towards_2023}, mobile phone usage patterns~\cite{blumenstock2015predicting}, nightlight intensity~\cite{keola2015monitoring}, and social media activity~\cite{asongu2021social}, which offer the potential for faster and more cost-effective poverty assessment at fine spatial resolutions. Beyond poverty mapping itself, such approaches are also increasingly important for evaluating whether development interventions and policies actually improve living conditions in relation to the SDGs. As noted by Daoud et al.~\cite{daoud_chinese_2026}, debates persist regarding whether development projects lead to measurable improvements in well-being, partly because observational estimates may be biased by incomplete adjustment and because reliable outcome data are often scarce at the neighbourhood level. Novel data sources and spatially detailed analytical approaches may therefore provide important new opportunities not only for identifying poverty, but also for monitoring and evaluating development outcomes over time.

Recent disruptions to major survey infrastructures, including the abrupt shutdown of the Demographic and Health Surveys (DHS) program administered by USAID, have further illustrated the fragility of conventional poverty data systems and the importance of developing complementary approaches for continuous monitoring and evaluation.

% \textbf{old}

% The use of unconventional data sources to assess and predict poverty has gained significant traction in recent years, driven by the need for timely, granular insights into economic well-being. This is highly relevant for tracking progress towards the Sustainable Development Goals (SDGs). Traditional poverty assessments, such as surveys and censuses, are often costly, time-consuming, and infrequent, particularly in developing countries where resource limitations and/or political constraints hinder comprehensive data collection \cite{burke_using_2021}. As a response, researchers have explored innovative data streams such as satellite imagery \cite{hersh_open_2021, sarmadi_towards_2023}, mobile phone usage patterns prediction \cite{blumenstock2015predicting}, nightlight intensity \cite{keola2015monitoring}, and social media activity \cite{asongu2021social}, which offer the potential for faster and more cost-effective poverty assessment at fine spatial resolutions.

% \textbf{old}

Out of the above, satellite imagery has received increased attention. The breakthroughs in the use of nighttime lights to observe global economic activity two decades ago have paved the way for more detailed and advanced observation \cite{chen2011using}. However, satellite imagery is inherently unstructured, requiring significant interpretation and processing to extract meaningful insights. This complexity can introduce challenges in analysis, but the availability of large volumes of high-quality satellite imagery combined with advancements in analytics, including machine learning and deep neural networks, has enabled even more powerful insights. 
Satellite imagery provides consistent coverage across diverse geographic areas and allows for monitoring changes over time, sometimes on an hourly scale. 
%Satellite imagery offers a feasible way to close some of the data poverty gaps that are so common around the globe. 
By integrating remote sensing and machine learning, researchers can address the data gaps faced by traditional methods, providing more timely and scalable insights into socioeconomic conditions. However, training custom neural networks for such analysis can be challenging due to the enormous data requirements and computational resources needed. The availability of sufficient labeled data is crucial for model training, and the complexity of interpreting unstructured data, such as satellite imagery, further complicates this process.

In this context, we explore a novel approach: leveraging Large Language Models (LLMs) to interpret and analyze satellite imagery to derive insights about village-level human poverty. LLMs, initially developed for natural language understanding, have shown remarkable adaptability across diverse domains, including image captioning, pattern recognition, and even geospatial analysis. However, their potential for extracting socioeconomic indicators directly from satellite images remains unexplored. Recent advancements in language models with vision capabilities provide an opportunity to extend traditional satellite-based analyses. Vision-enabled LLMs can leverage both pre-trained knowledge and visual data to analyze images in a broader context, reducing the need for extensive, domain-specific training.

Our work aims to assess the feasibility and accuracy of utilizing LLMs for poverty estimation using satellite data. 
Specifically, we address the question: Can LLMs effectively analyze satellite images to provide information related to human poverty? 
Our objective is to determine whether LLMs can provide \emph{reliable}, \emph{efficient}, and \emph{interpretable} insights at the village level, and to identify the opportunities and challenges of using such models for socioeconomic research. If LLMs can extract relevant information from satellite images as well as humans, then this would enable large-scale and potentially more interpretable poverty analysis from satellite images.
We find that by using a pairwise comparison approach, it is possible to get ChatGPT to rank satellite images based on poverty level, with results that are not significantly different from a Random Forest trained on human-expert-defined features. This analysis provides insights into the strengths and limitations of using LLMs for socioeconomic assessment and contributes to the ongoing efforts to utilize unconventional data sources for welfare analysis.

\subsection{Background}
In recent years, there has been a surge in the usage of LLMs for a myriad of applications. Given that these models are trained on almost all the freely available knowledge on the internet, they are quite attractive for exploratory studies. Additionally, Large Vision-Language Models (VLMs) have emerged quickly with impressive abilities in analyzing visual data and are constantly revised by the companies that create them. 
% For example, GPT-4V was released in 2023, GPT-4o, Claude Sunnet 3.5, and Gemini 2.0 were released in 2024, while GPT-5, Claude Sunnet 4.5, and Gemini 3.0 were introduced in 2025.

\subsubsection{Large Vision-Language Models}
Despite the fast-paced improvements, VLMs have not yet proven to be consistently reliable in analyzing Earth observations due to their weakness in spatial reasoning \cite{zhang_good_2024}. On the other hand, they have the advantage of transparency and explainability. For example, if one asks why a VLM thinks a specific satellite image is related to an area with high poverty, the model could provide a rationale for its decision-making even though its faithfulness is not guaranteed~\cite{turpin_language_2023}. This is a great advantage over purely vision-based deep learning models, which could not explain their decisions~\cite{hall_review_2022}. These characteristics are particularly relevant in development research contexts, where labeled training data are often scarce, transparency is important for trust and accountability, and analytical tools are frequently used to support exploratory reasoning rather than fully automated decision-making.

To overcome the problem with spatial reasoning, there have been efforts to specialize models only on Earth observations. GeoChat~\cite{kuckreja_geochatgrounded_2024} and EarthGPT~\cite{zhang_earthgpt_2024} are two of the most influential papers in this group. Nevertheless, it has been shown that specialized models are not better than general multimodal LLMs such as GPT-4o in geospatial tasks~\cite{danish_geobench-vlm_2025}.

On the other hand, newer versions of the commercial multimodal LLMs are being continually developed. GPT-5, Claude Sonnet 4.5, and Gemini 3.0, all introduced in 2025, are among this group. Notwithstanding, the newer models do not seem to be much better than the previous ones. For example, GPT-4o is still competitive compared to GPT-5 in visual tasks~\cite{openai_introducing_2025}.

\subsubsection{Ranking from Comparisons}

It has been demonstrated that LLMs, even when trained using the correct pairwise preferences, cannot accurately assign higher probabilities to better answers, which is an issue with current preference training methods \cite{chen_preference_2024}. On the other hand, it has been shown by \cite{qin_large_2024} that taking advantage of Pairwise Ranking Prompting (PRP), they were able to get better results than other point-wise and list-wise ranking methods. Nevertheless, their approach is for ranking text documents, and pairwise comparisons are converted to ranking via a naive bubble-sort-like sliding window algorithm. This might not work well in handling the noise in the LLM's decisions, which is a natural characteristic of a deep generative model. Nevertheless, relative ranking approaches are well-aligned with development practice, where decisions often involve prioritizing locations or populations under constrained resources, rather than estimating precise absolute values.

% Efficient sorting algorithms, such as quicksort and mergesort, could be used as an effective sampling strategy in ranking from pairwise comparisons \cite{maystre_just_2017}.

In this paper, we propose addressing the limitations of current methods in spatial reasoning and poverty-ranking by combining pairwise prompting of a VLM with state-of-the-art ranking algorithms based on pairwise comparisons. These methods take advantage of the well-tested Bradley-Terry-Luce model~\cite{NegahbanOS2012}, which is a probabilistic approach tolerant to noise. From this family, we have used three algorithms with publicly available implementations consisting of Rank Centrality~\cite{negahban_rank_2017}, Luce Spectral Ranking, and Iterative Luce Spectral Ranking~\cite{maystre_fast_2015}.
% \subsection{Quality of Ground Truth Data}
% Public poverty index datasets such as the one from~\cite{DHS-Keys-VII} are anonymized with additive location noise. Such measures, despite the alleviation of privacy concerns, reduce the quality of data for training and evaluation purposes. X~\cite{burke_using_2021}
From a development perspective, this approach offers a transparent and noise-tolerant way to aggregate uncertain judgments, which is particularly valuable when ground-truth socioeconomic data are imperfect or contested.

\subsubsection{Deployment Ethics and Explainability}

In the ICT4D literature, it has been argued that AI-driven solutions should not only emphasize human agency but also help to create positive economic cycles. Additionally, they should not increase inequality by predicting and modifying human behavior~\cite{qureshi_cycles_2023}. AI-driven solutions that estimate poverty from satellite images could augment costly surveys and keep up with fast changes in human poverty. The explainability of the AI model is quite important to bring back agency to humans in decision-making. Quick and effective detection of poverty could enable the prevention of negative economic cycles. The findings of~\cite{hall_review_2022} regarding the lack of explainable models show an urgent need for explainable poverty estimation models based in the remote sensing domain. 

Even the best algorithmic solutions should not be deployed blindly. Purely focusing on statistical parities might backfire in closing inequality gaps. Correct policies need to be put in place to ensure a positive effect on the fair distribution of social goods~\cite{zezulka_fair_2024}. On a technical level, any AI algorithm needs to be monitored for biases and to actively mitigate privacy risks that could come with its predictions~\cite{ghamisi_responsible_2025} (e.g., stigmatization of communities by public release of the predicted poverty scores).

Our ranking from the pairwise comparisons method based on the outputs of a VLM allows for a model where each comparison could accompany a textual explanation and is influenced by modifying the prompt. This transparency and flexibility in deployment enable easier discovery and mitigation of biases, along with monitoring and enforcement of policies such as social equality and privacy. We should mention that, in this paper, we did not collect or evaluate the textual rationale of the model for its decisions. This is a task left to be done in future studies.

\section{Results}

\begin{figure*}[t]
    \includegraphics[width=\textwidth]{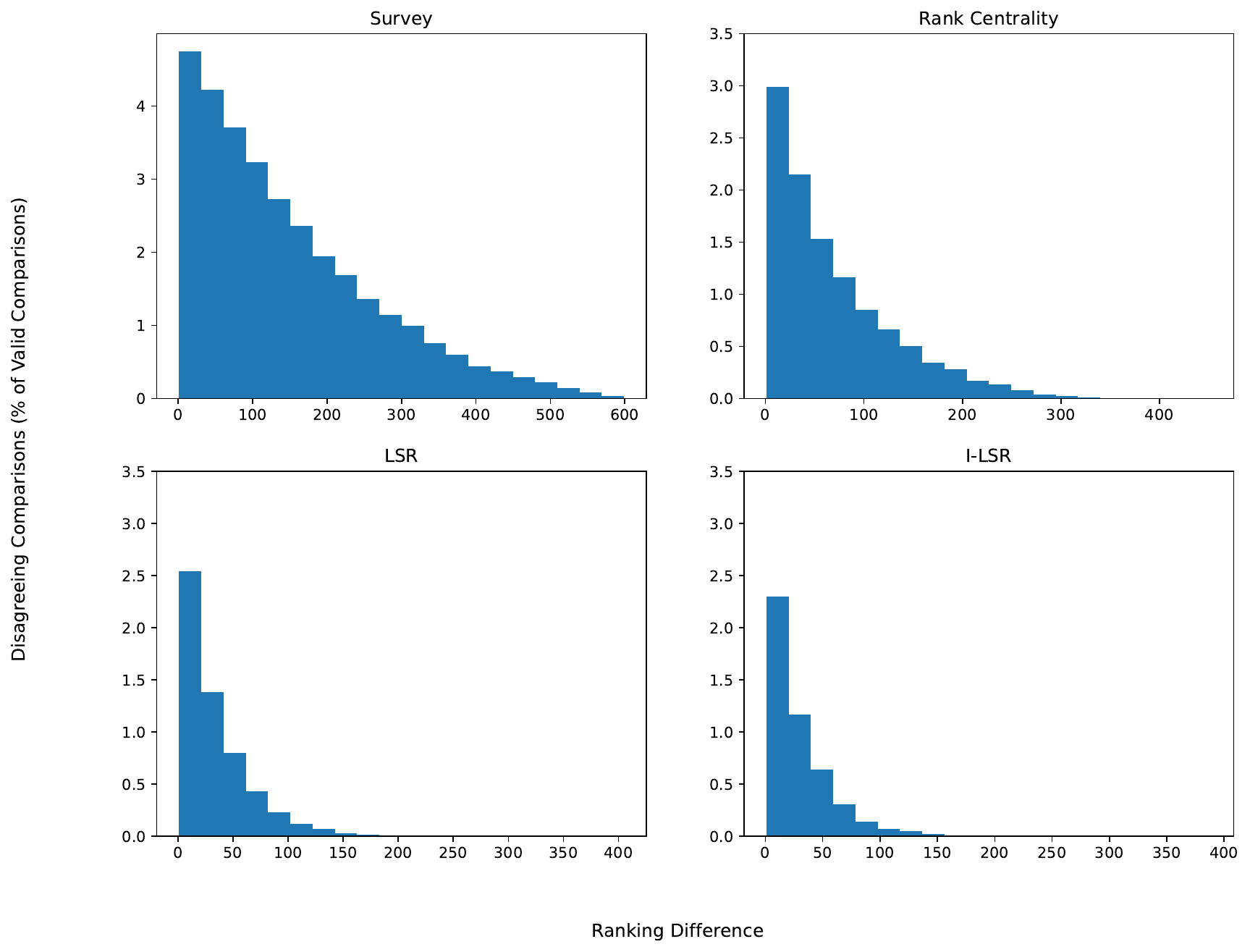}
    \caption{Upper left panel: Histogram of the differences in HV271 rankings for the pairs where the ChatGPT pairwise comparison disagrees with what would be expected from HV271 (i.e., when the wealthier/poorer comparison has opposite signs in the HV271 and the ChatGPT comparison). Upper right and lower panels: Histograms of the differences in rankings between pairs in the final ChatGPT ranking and the pairwise comparison matrix; the panels show the results for the three algorithms tested to infer the overall ranking from the pairwise comparison matrix. Compare with numbers in Table~\ref{tab:ranking_inference_results}.}
    \label{fig:histograms_rank_diffs}
\end{figure*}

\subsection{Developing the prompt}

The final prompt presented in Section~\ref{sec:pairwise_prompt} was developed through a sequence of experimental steps.

Initially, we experimented with more loosely formulated prompts, such as “Which image is wealthier?” However, the approach proved unreliable when using such high-level concepts effectively without detailed instructions. The lack of specificity in these prompts led to inconsistent or ambiguous results, highlighting the need for clear and measurable criteria to guide the analysis.

The goal was to reach a reliable method to assess and compare the relative visual wealth of two images using observable indicators tied to economic or developmental status. The analysis required clear and actionable results, expressed as a binary conclusion (“Image 1 is wealthier” or “Image 2 is wealthier”) to ensure practical utility and consistency.

To achieve this, we formulated key indicators of wealth that are visually measurable and relevant across different contexts. These included: the quality of infrastructure, reflecting economic investment and maintenance; the number of floors in buildings, indicative of development and resource allocation; the presence of visible greenery or well-maintained areas, a marker of environmental care often associated with higher economic status; and visible amenities such as paved roads and power lines, which signify advanced public infrastructure and services.

While this approach draws inspiration from methods like the DHS asset wealth index, which evaluates household wealth based on observable assets and infrastructure, it is not identical. Our method focuses on visual indicators observable in images, emphasizing comparative assessment rather than deriving an absolute wealth score.

Notably, OpenAI's models initially faced restrictions on engaging with topics like wealth and welfare due to the complexity and sensitivity of these subjects. Over time, advancements in model capabilities and refinements in guidance allowed for more nuanced discussions grounded in structured frameworks. By 2024, the OpenAI model GPT-4o (released May 2024) demonstrated improved performance in analyzing socioeconomic indicators when given clear, detailed instructions. This development enabled the formulation of prompts like ours, which rely on specific, observable criteria to achieve reliable results. The final prompt was designed to explicitly list these indicators, ensuring clarity and focus in the analysis while maintaining a scalable approach. 

\subsection{From pairwise comparisons to the overall ranking}

%\textcolor{blue}{The Bradley-Terry-Luce model is a well-established approach to extract ranking from pairwise comparisons.}

The number of images is 608, and the number of pairwise comparisons is ${608 \choose 2} = 184,528$. Out of these, 26,757 were unresolved after two attempts and excluded.
The remaining 157,771 were used to construct the final ranking with the I-LSR algorithm. We denote these as \emph{valid} comparisons.

The consistency of the 157,771 comparisons was checked by creating a directed graph from the comparisons and counting the number of 3-cycles in this graph. A 3-cycle corresponds to three sites that do not follow the transitivity rule, i.e., they behave like Penrose stairs (A is wealthier than B, which is wealthier than C, which in turn is wealthier than A). We found 155,546 such 3-cycles. 
To get a scale for the number of 3-cycles, we can assume that all edges belong to a clique, which gives us $n\approx \sqrt{2E}$, where $n$ is the number of nodes in the clique and $E$ is the number of edges (comparisons). With $E = 157,771$, we have $n \approx 562$. The maximum number of 3-cycles of a tournament graph with $n$ nodes is at least $n(n^2-4)/24$~\cite{grzesik_cycles_2023}, which gives 7,395,920 possible 3-cycles. Hence, the number of observed cycles (155,546) is only about 2\% of the maximum number of possible 3-cycles. This indicates a low rate of contradictions between the pairwise comparisons.

The three methods for inferring the total rank inference (I-LSR, LSR, and RC) were tested, and the number of conflicts and the total rank disagreement for their solutions are shown in Table~\ref{tab:ranking_inference_results}. The number of conflicts is simply the number of pairwise comparisons that disagree with the final ranking.
The total rank disagreement is the sum of absolute rank differences between pairs whose pairwise comparison disagrees with the final overall ranking. This is illustrated in the top right and two bottom subfigures in Fig.~\ref{fig:histograms_rank_diffs}; the conflicts are just the sum over the bins, and the total rank disagreement is the sum of the bins in the histograms weighted by their bin value.
A solution with a large total rank disagreement disagrees more with the pairwise comparison table than a solution with a smaller total rank disagreement.
The pairwise comparison matrix is quite dense (many pairwise comparisons were done), so we used the lowest regularization parameter ($\alpha$) possible; a value of $\alpha = 0$ caused errors. Table~\ref{tab:ranking_inference_results} shows that the I-LSR algorithm gave somewhat better results, and they were used in the further analysis.

\begin{table}
    \centering
    \begin{tabular}{c c c c} \toprule
        &  $\log_{10}(\alpha)$ & Number of conflicts & Total rank disagreement \\
        \midrule
        I-LSR & -297 & \textbf{7,483} & \textbf{226,388}\\ 
        LSR & -306 & 8,923 & 308,000 \\
        RC & -323 & 17,240 & 1,191,102\\
       \bottomrule
    \end{tabular}
    \caption{Number of conflicts and total rank disagreement for the three algorithms solving the Bradley-Terry model. The $\alpha$ is the regularization parameter we used for these algorithms in the choix Python library. The boldface marks the best result.}
    \label{tab:ranking_inference_results}
\end{table}

\subsection{The quality of the rankings}

\begin{figure*}[t]
    \centering
    \subcaptionbox{CNN 3x3 ($\rho = 0.78$)\label{fig:CNN3x3_rank}}{\includegraphics[width=.4\textwidth]{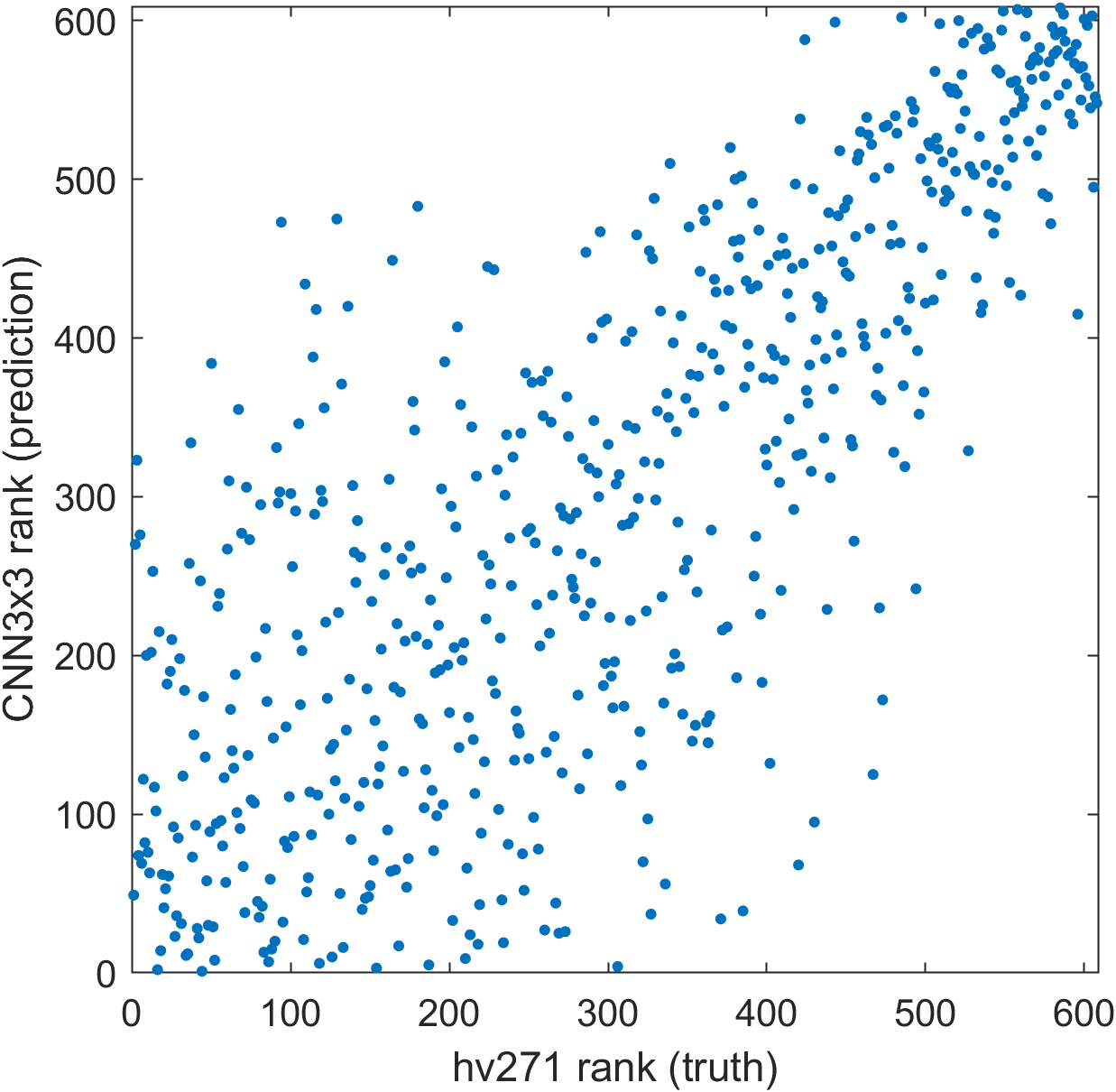}}
    \subcaptionbox{ChatGPT ($\rho = 0.56$)\label{fig:ChatGPT_rank}}{\includegraphics[width=.4\textwidth]{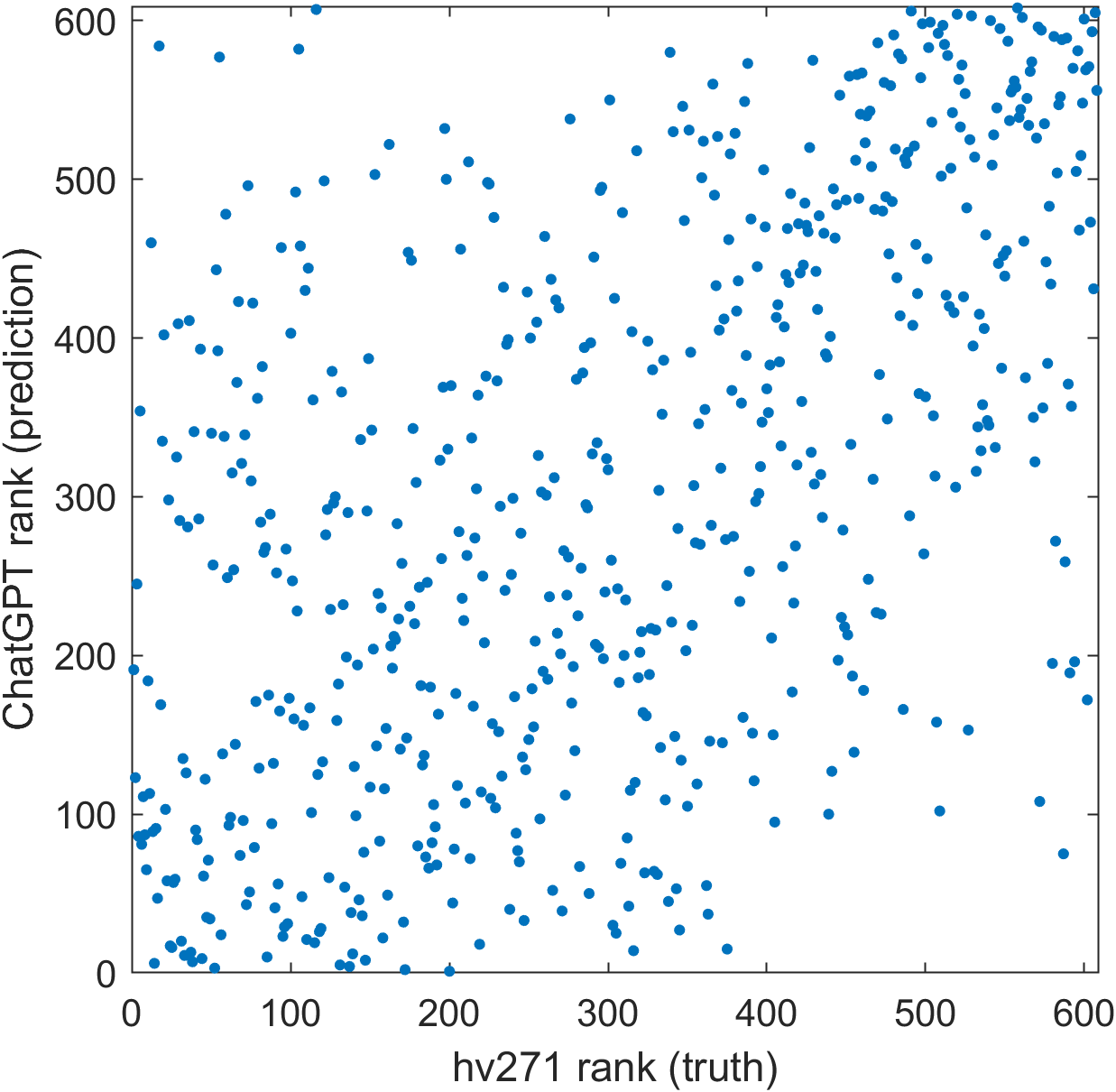}}
    
    \caption{Scatter plots showing the ranked wealth predictions versus the ranked true wealth (averages of HV271) for the 608 DHS clusters. The $\rho$ value is Spearman's rank correlation.}
    \label{fig:rank_plots}
\end{figure*}

\begin{sidewaysfigure*}
    \centering
    \includegraphics[width=\textwidth]{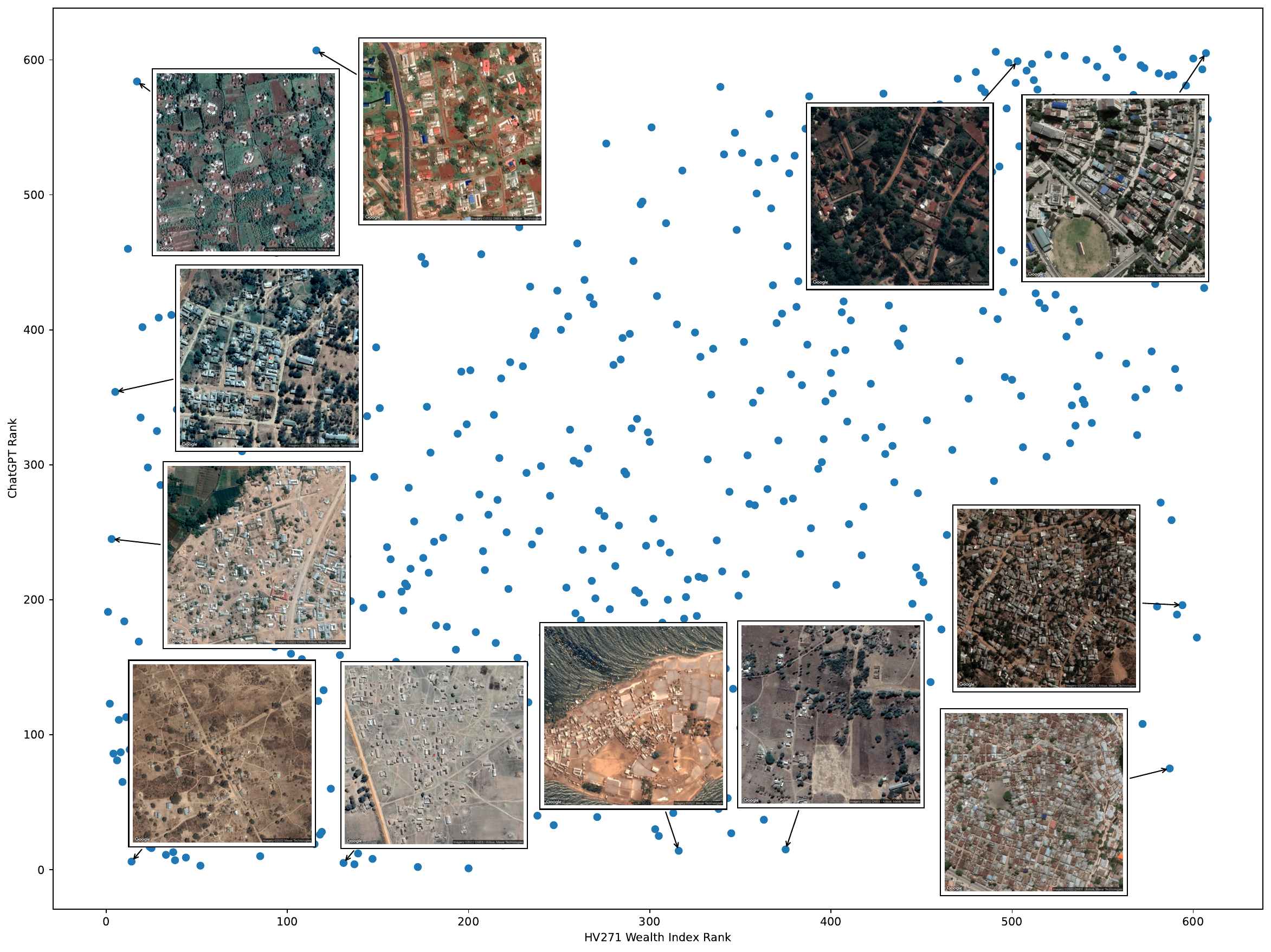}
    \caption{Examples of images related to different points where the ranking obtained from ChatGPT disagrees with the one from the HV271 wealth index.}
    \label{fig:rank_plots_figs}
\end{sidewaysfigure*}

The upper left panel in Figure~\ref{fig:histograms_rank_diffs} illustrates the agreement between the pairwise comparisons made by ChatGPT and a ranking by the DHS HV271 value. The x-axis shows the difference in rank, when ranked by HV271, between two sites when the ChatGPT pairwise ranking disagrees with the ranking by HV271. 

The ranking from the ChatGPT pairwise comparisons is an overall wealth ranking of the DHS sites, not the actual wealth index value. All evaluations are therefore based on ranking or grouping performance. The ChatGPT rankings are compared to a CNN-based estimation and a Random Forest estimator that uses image features that have been coded by domain experts. The CNN reflects the state-of-the-art in estimating wealth from satellite images. The Random Forest reflects the performance of a domain expert in detecting relevant features and wealth estimation from this.

Figure~\ref{fig:rank_plots} illustrates ChatGPT's ranking performance and the ranking performance of the CNN model. The Spearman's rank correlation is $\rho = 0.78$ for the CNN and $\rho = 0.56$ for the ChatGPT rank. From a development perspective, this level of agreement suggests that our approach, while being far more explainable, may be suitable for coarse-grained prioritization or screening tasks rather than precise poverty estimation.

If the predicted wealth index values are used for the CNN and compared to the actual wealth index, then the Pearson correlation is $r = 0.85$, and the coefficient of determination is $R^2 = 0.72$, which compares favorably to those reported by Jean et al.~\cite{JeanEtAl2016} for Tanzania (in the Supplementary material). Figure \ref{fig:rank_plots_figs} shows the scatter plot for the ChatGPT ranking versus the ranked true wealth, the same as in Figure \ref{fig:rank_plots}, with added images for a small selection of ranking data points. The selection represents images where ChatGPT agrees well with the HV271 ranking (lower left and top right), but also where they fully disagree (top left and lower right).

\begin{table}
    
    \centering
    \begin{tabular}{l c c c c}\toprule
    
         & \multicolumn{2}{c}{Rank Correlation} & \multicolumn{2}{c}{Quintile MCC}  \\
         \midrule
         CNN $3\times3$ & {\bf 0.78} & [0.75 0.82] & {\bf 0.38} & [0.33 0.43]\\
         Random Forest & 0.59 & [0.53 0.64] & 0.20 &[0.16 0.25]\\
         ChatGPT & 0.56 & [0.50 0.61] & 0.20 & [0.15 0.25]\\
         \midrule
         V191 & 0.99 & [0.99 0.99] & 0.86  & [0.82 0.89]\\
         Average Random Guess & 0.00 & [-0.06 0.06] & 0.00 & [-0.03 0.03] \\ 
         \bottomrule
    \end{tabular}
    \caption{Evaluation of regressor models' predictions compared to the ground truth wealth index. The numbers in bold are the significantly best prediction results for each evaluation metric. The 95\% confidence interval via bootstrapping is also presented for each measured value. Results for the V191 wealth index and average random guess are shown for reference.}
    \label{tab:rank_corr_and_mcc}
\end{table}

Table~\ref{tab:rank_corr_and_mcc} shows the Spearman's rank correlation and the quintile MCC for the ChatGPT rank, comparing it to the two reference models: the CNN and the Random Forest. It is clear that the CNN-based wealth prediction is the best. However, the ChatGPT ranking is not significantly different ($p < 0.05$) from the Random Forest model in the satellite images when calculating the ranking correlation and quintile MCC, as shown in Table~\ref{tab:randomforset-chtgpt}. The ChatGPT and the Random Forest results are not significantly different. The two bottom rows are provided for reference. The ``V191'' row shows the correlation and match between two true values: wealth estimated at the household and individual level (see Section~\ref{DHSinfo}), and presents how the target would change with an alternative aggregation method (household vs. women-weighted) and is not a ceiling on performance. The ``Random guess'' row shows the average performance when 10,000 random guesses are made on the wealth index for each DHS cluster.

Table~\ref{tab:Dichotomies_MCC_hv271} shows a more detailed performance comparison, where we explore the ChatGPT performance when grouping DHS clusters into two groups. First, we separate the poorest quintile from the others, denoted (1 vs. 2-5); then we separate the two poorest quintiles from the others, denoted (1-2 vs. 3-5), and so on. 
The Random Forest result for (1 vs. 2-5) is not significantly better than randomly guessing. ChatGPT is significantly better than Random Forest for (1 vs. 2-5), and Random Forest is significantly better than ChatGPT for (1-4 vs. 5). For the other cases, the ChatGPT and Random Forest results are not significantly different. The two bottom rows are provided for reference. The ``V191'' row shows the match between two true values: wealth estimated on the household and on the individual level (see Section \ref{DHSinfo}), and represents an upper limit for the result. The ``Random guess'' row shows the average performance when 10,000 random guesses are made on the wealth index for each DHS cluster.
The CNN reference model is still the best, and ChatGPT and the Random Forest models are very similar. However, there is a small difference between ChatGPT and the Random Forest model; ChatGPT appears to be better at detecting the very poor, whereas the Random Forest model is better at detecting the most wealthy (1-4 vs. 5). We have also studied the significance ($p < 0.05$) of the difference between Random Forest and ChatGPT in binary MCC \emph{over dichotomies}. The results are shown in Table~\ref{tab:randomforset-chtgpt}, where it can be seen that 3 out of 4 dichotomies are significantly different, and only the 1-3 vs. 4-5 dichotomy is not.

\subsection{Analysis of the Comparison Quality}
The repository accompanying this paper contains data of the comparisons and their results in the same order they were performed.
We computed the median of the rolling average on the resolving state of the comparisons in the original order they were executed. With a window size of 1000, the median was computed to be $14.50\%$, which is exactly the same as the overall proportion of comparisons that stayed unresolved after two retries. On the other hand, the minimum unresolved comparisons per image was 3, and the maximum was 333, with a median of 77 and a 95th percentile of 196. Given that the execution of pair comparisons was done in random order, these quantities show that unresolved comparisons are image-specific and not related to a period of trouble with the API.

The correlation between the wealth index of the images and the unresolved state of their comparisons was also calculated. The Pearson $\rho$ was 0.13 with a p-value of 0.002, showing a weak positive correlation.

To check the robustness of the comparison decisions to the order of images, we counted the number of times that the first image was chosen as wealthier versus the times the second image was wealthier in the resolved comparisons. These numbers are $51\%$ and $49\%$, respectively, which have a p-value equal to $1.5\times10^{-22}$ using a binomial test. Hence, the order of the images in the comparisons has no significance ($p < 0.05$) on the result.

\begin{table*}

    \centering
    \begin{tabular}{l c c c c c c c c}\toprule
         & \multicolumn{8}{c}{Binary MCC over quintile groups} \\ 
         \cmidrule{2-9}
         &  \multicolumn{2}{c}{1 vs. 2-5}& \multicolumn{2}{c}{1-2 vs. 3-5}& \multicolumn{2}{c}{1-3 vs. 4-5}& \multicolumn{2}{c}{1-4 vs. 5}\\ \midrule
         CNN $3\times3$ & {\bf 0.39} & [0.30 0.48] & {\bf 0.53} & [0.46 0.59] & {\bf 0.73} & [0.67 0.78] & {\bf 0.74} & [0.67 0.81]\\
         Random Forest & 0.08 & [0.00 0.17] & 0.44 & [0.36 0.51] & 0.53 & [0.46 0.60] & 0.53 & [0.44 0.61]\\
         ChatGPT & 0.26 & [0.17 0.35] & 0.33 & [0.25 0.40] & 0.48 & [0.41 0.55] & 0.42 & [0.33 0.50]\\
         \midrule
         V191 & 0.89 & [0.84 0.93] & 0.92 & [0.88 0.95] & 0.95 & [0.93 0.98] & 0.95 & [0.91 0.98]\\
         Average Random Guess &0.00 & [-0.06 0.06] & 0.00 & [-0.06 0.06] & 0.00 & [-0.06 0.06] & 0.00  & [-0.06 0.06]\\
         \bottomrule
    \end{tabular}
    \caption{Binary MCC values for the four possible dichotomies on the wealth quintiles. The numbers in bold indicate the significantly best prediction result for each dichotomy. The 95\% confidence interval via bootstrapping is also presented for each measured value. Results for the V191 wealth index and average random guess are shown for reference.}
    \label{tab:Dichotomies_MCC_hv271}
\end{table*}

\begin{table*}

    \centering
    {\begin{tabular}{l c c c c c c}\toprule
          \multirow{2}{*}{}&\multirow{2}{*}{Rank Correlation} &  \multirow{2}{*}{Quintile MCC} &\multicolumn{4}{c}{Binary MCC over quintile groups}\\ 
         \cmidrule{4-7}
         & & & 1 vs. 2-5& 1-2 vs. 3-5& 1-3 vs. 4-5& 1-4 vs. 5\\ \midrule%\cmidrule{2-7}
         95\% CI& [-0.03 0.09] &  [-0.05 0.07] &  [-0.28 -0.07] & [0.02 0.18] & [-0.04 0.13] &  [0.01 0.22] \\
         Significantly different? & No & No & Yes & Yes & No & Yes \\
         \bottomrule
    \end{tabular}}
    \caption{The confidence intervals (CI) of the difference between Random Forest and ChatGPT ($Random Forest - ChatGPT$) and the significance of the two predictors being different ($p < 0.05$) for the reported performance metrics. The confidence intervals are obtained via bootstrapping on the 608 clusters.}
    \label{tab:randomforset-chtgpt}

\end{table*}

\section{Discussion}

This study explored the potential of utilizing LLMs for village-level human poverty predictions using satellite images. Specifically, the GPT-4o model by OpenAI was instructed to rank pairs of images, concluding which image shows a wealthier location than the other. All possible pairwise comparisons were done and combined into a final wealth ranking of all images. The results show that the ability of the LLM is comparable to a supervised model whose inputs are human-identified image features, which in turn is better than if domain experts are asked to estimate wealth directly from the image, according to a previous study~\cite{sarmadi_human_2024} in which domain experts estimated the wealth level directly from images similar to those presented to the LLM, and it was compared to a Random Forest model trained on human-derived features. It was concluded that human ranking was inferior to the Random Forest model. The very same model is compared in the current manuscript to the LLM's ability to rank images, and as shown by the results, the two approaches are comparable in performance. In the current study, we did not ask experts to perform pairwise ranking themselves. Another limitation is that our prompts requested a one-phrase answer, and if the answer mentions both images, it is discarded.

We find the aforementioned results notable in several aspects. The domain experts have several years of regional experience and high educational levels. The LLM used in this study possesses strong language skills and outperforms humans in many aspects, but was not fine-tuned for the particular task of estimating wealth in Tanzania. Even though the prompt was designed to include key indicators of wealth that should be visible in the images, it lacked other possibly helpful factors, such as scale and regional information. Furthermore, the performance could most likely be increased by an elaborate chain-of-thought prompting, where examples of reasoning steps would help the LLM reach a better ranking. The LLM approach is also scalable. The pairwise comparison method employed in this study allowed for rapid analysis of a large dataset (608 sites and nearly 200,000 comparisons). This efficiency contrasts sharply with extensive manual annotations performed by domain experts, which are resource-intensive and time-consuming. The LLM ranking of the 608 images was performed by comparing all possible pairs of images. A more selective approach can likely be employed to reduce the number of pairwise comparisons, still achieving a final accurate ranking of the images. While the overall performance of the GPT-4o model was similar to the Random Forest model, the more granular comparison presented in Table~\ref{tab:Dichotomies_MCC_hv271} indicates that the LLM has better performance for the images representing very poor areas. One can speculate why. Perhaps the text corpus and the image collections seen by the LLM have a bias towards characteristics of poor areas rather than non-poor ones. 

The CNN approach is clearly very capable of estimating wealth from the satellite image, and there have been further developments on such models. The highest accuracies with a CNN have been achieved by Lee \& Braithwaite \cite{LeeB22} who, e.g., reported $R^2 = 0.90$ for Tanzania. We did not compare to their specific approach since our purpose was to illustrate the remarkable ability of ChatGPT, not to present the best wealth prediction. Also, the code for the results in \cite{LeeB22} is not available, and they make some hard-to-replicate adjustments to the data, which makes it difficult for us to reproduce their model. However, the CNN approach used here for comparison is well-tested, which is why we chose this as a benchmark.

Figure \ref{fig:rank_plots_figs} presents three interesting observations regarding the relationship between ChatGPT-derived rankings and the HV271 wealth index rankings (considered the ground truth).

Strong agreement between rankings: In several instances, the rankings provided by ChatGPT align closely with the HV271 wealth index rankings. These cases highlight the model's ability to effectively analyze image features and identify socioeconomic indicators. Poor images are characterized by sparsely populated areas, dryness, and a lack of greenery, which align with our understanding of poverty. Conversely, wealthy images typically feature green vegetation and town- or city-like spatial layouts, which are consistent with our expectations. Notably, ChatGPT rankings align well with subjective appraisals, supporting their plausibility.

False positive wealth assessments: Some images are ranked as wealthy by ChatGPT, despite being classified as poor according to the HV271 wealth index. These images share similarities with those ranked as wealthy and have little in common with those ranked as poor. Observing the images along the y-axis reveals that ChatGPT’s rankings appear reasonable, even if they diverge from the ground truth.

False negative wealth assessments: Conversely, there are cases where ChatGPT ranks images as poor, while the HV271 wealth index identifies them as wealthy. The images along the x-axis are not as separable or easy to understand. These wealthy images differ from those ranked wealthy by both ChatGPT and the HV271 index. While they are still high-density, they exhibit a different, small-scale spatial arrangement and minimal greenery.

These observations suggest that ChatGPT’s image-based socioeconomic analysis follows a logical pattern that aligns well with subjective appraisals based on visual cues. For the images analyzed, ChatGPT appears to identify plausible indicators of wealth or poverty, supporting its reliability in this context. Such capabilities may be particularly relevant for development researchers, NGOs, or statistical offices seeking exploratory tools to complement existing data collection rather than replace established survey infrastructures. However, discrepancies with the HV271 wealth index raise the possibility that issues in the ground truth data, such as displaced DHS coordinates or limitations in the index itself, could contribute to the observed misalignments. Further investigation is needed to determine the extent to which these factors influence the discrepancies.
{%\color{Blue}

A natural follow-up to our work would be the expansion of the validation regions beyond Tanzania and verifying our findings across multiple regions in Africa, while also validating the temporal correlation with the ground truth wealth indices. This type of undertaking needs more time and resources; however, it is the gold standard for validation in this field set by~\cite{yeh_using_2020}.

Our paper has taken advantage of GPT-4o for pairwise comparisons. While this multimodal frontier model is considered powerful to this day, newer VLMs are constantly being introduced. We believe that our analysis needs to be reevaluated with the newer frontier models in the future to keep up with the innovations in this area. 

On the other hand, there are VLMs already specialized for remote sensing, such as the works of \cite{zhang_earthgpt_2024} and  \cite{kuckreja_geochatgrounded_2024}. While they normally cannot compete with the frontier models such as GPT-4o~\cite{danish_geobench-vlm_2025}, there are promising efforts to create VLMs with better geospatial reasoning~\cite{xu_geo-r1_2025}. This type of model is the next candidate to be combined with our pairwise comparison ranking method.

}

We found the performance of ChatGPT-4o quite remarkable in ranking the areas based on their wealth level using remote sensing images without any pretraining. It should be mentioned that in our prompt, we did not give any information regarding the area the images correspond to. This includes the scale of the area covered by the region and any information about its whereabouts. CNNs proved to perform better, however, by taking advantage of supervised learning. On the other hand, the visual analysis of disagreements between ChatGPT's ranking and the ground truth gave us the impression that it could either be caused by anonymizing displacement~\cite{burgert_geographic_2013} in the coordinates of the clusters or the visual changes in the satellite imagery since the 2015/2016 DHS survey.

Nevertheless, we believe that the results using the ChatGPT-4o model could be improved by enhancing the prompts via giving more contextual information about the images or prompt engineering techniques like providing a chain of thought for reasoning. Rather than replacing established poverty measurement practices, multimodal LLMs may expand the analytical capacity of development research in contexts where traditional data collection remains constrained, provided they are deployed transparently and responsibly.

\section{Methods}

%\subsection{Introduction}
%This study explores the use of vision-enabled language models to analyze poverty and welfare in Tanzania using satellite imagery. Our approach combines multiple data sources, including survey data from the Demographic and Health Surveys (DHS), satellite images, expert ratings, and CNN assessments. By integrating these diverse datasets, we aim to estimate and validate poverty indicators in a comprehensive manner, utilizing both qualitative and quantitative perspectives.

\subsection{Survey Data (DHS Wealth Index)}
\label{DHSinfo}

The ground truth data for this study
%The primary dataset for this study includes asset wealth index data for 608 georeferenced clusters.
%, which approximately represents a sample of towns or villages across Tanzania. 
%The data 
comes from the 2015/2016 Tanzania Demographic and Health Survey (DHS) dataset, which is based on 12,563 households, grouped into 608 clusters across the 30 regions of the country. Each cluster ranges in sample size from 12 to 22 households per cluster.

%This dataset acts as ground truth, as in many similar studies. 
%It serves as a benchmark against which other data sources can be compared and validated. 
While it is not without problems, such as potential inaccuracies and limitations in capturing all aspects of poverty, it remains the best available source of comprehensive socioeconomic information. 
In this study, the DHS wealth index is therefore treated not as an objective ground truth, but as the best available institutional approximation of socioeconomic status, subject to known sources of noise and uncertainty.

The DHS data is collected through extensive household surveys and provides critical socioeconomic indicators, including the asset wealth index. This index is a composite measure of a household's ownership of various assets, ranging from consumer goods to housing characteristics, and serves as a proxy for socioeconomic status. 

%The DHS wealth index is constructed based on household asset ownership, housing characteristics, and access to services, providing a composite measure of relative wealth. Despite its limitations, the DHS wealth index remains a widely used benchmark in poverty analysis. It comes with geospatial coordinates, enabling researchers to link the wealth index to other types of data, such as satellite imagery.

We base our ground truth on the DHS HV271 variable, \emph{the wealth index factor score}~\cite{DHS-Keys-VII}, which is calculated using principal component analysis on data on ownership of basic assets (e.g., bicycles, television sets, sanitation assets, and water access). It represents wealth on the household level. For each cluster, we use the average of the HV271 variable for all households in that cluster.

We also use the DHS V191 variable, which is the household-level wealth index (HV271) attached to individual women~\cite{DHS-Keys-VII}. Therefore, in the cluster, households are weighted by the number of eligible women, and households without women are eliminated. The value should have a large overlap with the DHS HV271 value, but V191 and HV271 are not identical, and we use the match between them to estimate the maximum possible accuracy in ranking by the wealth index. The wealth index is used as a proxy for poverty, as in similar studies. 

A recurring problem in poverty analysis is how to work with institutional data sources that serve as imperfect proxies for underlying socioeconomic conditions, which is essential for training and validation of models. The work from Burke et. al.~\cite{burke_using_2021} highlights the critical issue of noise in ground-truth data and its implications for machine learning models. Ground-truth data, such as survey information, often suffer from measurement errors, sampling biases, and temporal misalignment with satellite imagery. Additionally, some noise is introduced intentionally through techniques like jittering, which adds small random perturbations to preserve the privacy and integrity of respondents. These inaccuracies and intentional perturbations introduce significant challenges in the training and validation of machine learning models, particularly in contexts where reliable data is scarce. Temporal mismatches exacerbate these issues by misaligning features in satellite data with outdated or inconsistent ground observations, undermining the development of accurate predictive models. As a result, noisy ground-truth data can obscure important relationships, mislead model performance evaluations, and diminish the reliability of predictions.

Supervised machine learning models are particularly vulnerable to these issues because they rely heavily on the quality of their training data. Noisy labels can distort feature-label relationships, leading to misclassifications and reduced generalization across different spatial or temporal contexts. Models trained in one region and applied to another may fail due to regional variability and non-uniform data quality. Similarly, hybrid models combining satellite imagery with auxiliary data like socioeconomic surveys often struggle with integration when ground-truth datasets vary in quality or coverage. Even unsupervised or semi-supervised approaches, which are less dependent on labeled data, can be undermined by noise in preprocessing or validation stages, complicating the interpretation of results and the robustness of clustering or feature extraction efforts.

\subsection{Satellite Imagery}

\begin{figure*}[t]
    \centering
    \includegraphics[width=0.7\linewidth]{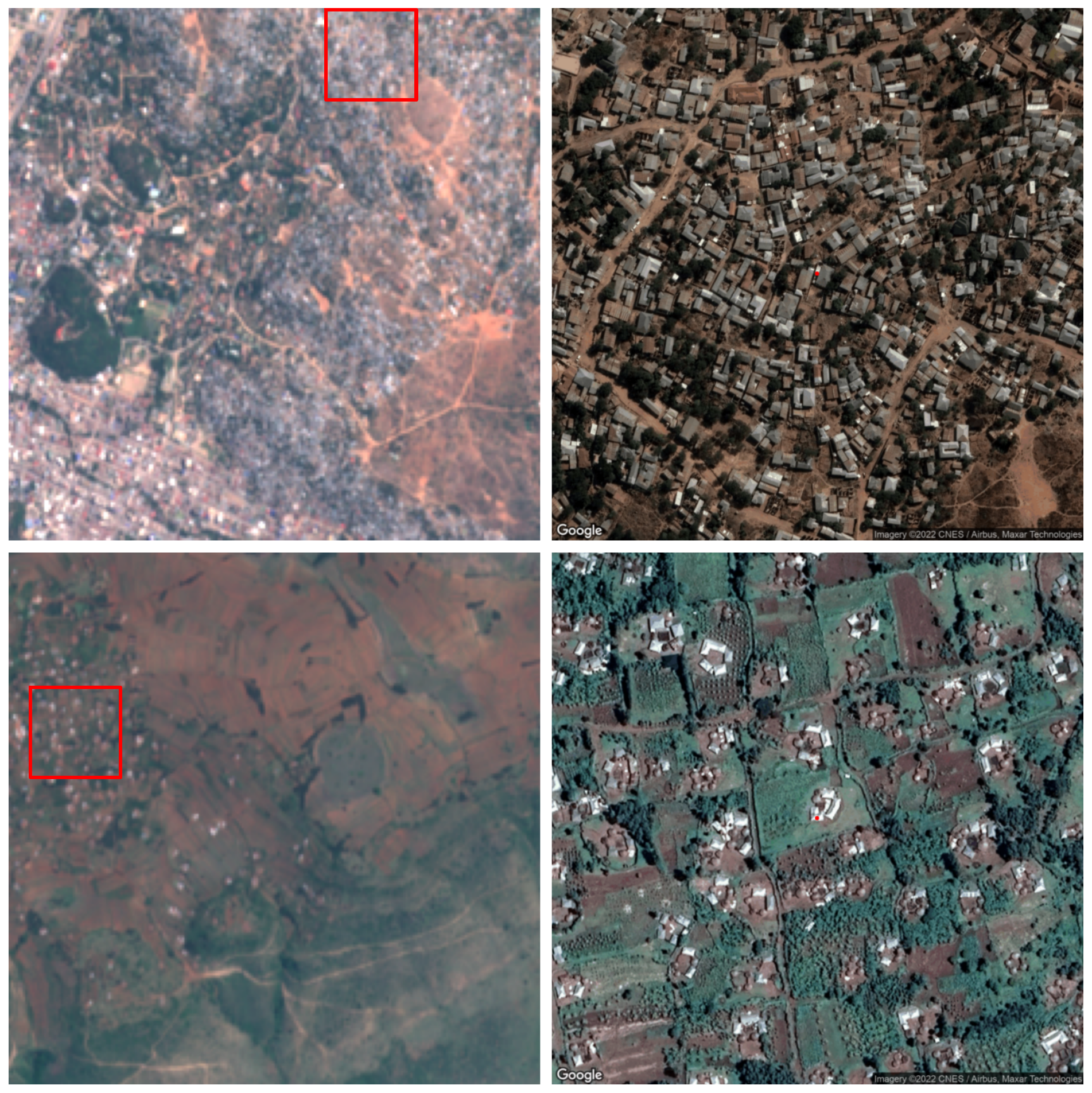}
    \caption{Examples of medium resolution images ($2 \times 2$ km) used in training the CNN (left) compared to their corresponding high resolution images shown to ChatGPT (right). The area for the high-resolution image is marked with a red square in the corresponding medium-resolution image.}
    \label{fig:enter-label}
\end{figure*}

Sentinel-2 imagery was used in the training of machine learning models based on convolutional neural networks, described below, while high-resolution Google imagery was used as input for LLM. These images were selected to closely match the spatial area of the surveyed clusters, allowing for a direct visual representation of the physical environment where the households reside. The satellite imagery, including high-resolution Google imagery and multi-spectral Sentinel-2 data, provides a rich source of information for understanding environmental factors such as building density, road infrastructure, vegetation cover, and land use patterns, which are often correlated with socioeconomic conditions. Figure \ref{fig:enter-label} shows example images of the same locations, from Google and from Sentinel-2. Together, these publicly available medium-resolution imagery and widely accessible high-resolution sources make the approach easier to replicate and more feasible to deploy for development researchers and institutions with limited computational resources.

%{\bf We need examples here of Google and Sentinel images of the same location}

%\subsection{Domain experts}

%The poverty levels of the locations for the satellite images were estimated by 102 domain experts with significant experience, half of them with more than 10 years of fieldwork experience in Sub-Saharan Africa. Their background is described in detail in Sarmadi et al.~\cite{sarmadi_human_2024}.

\subsection{Pairwise comparison of images}
\label{sec:pairwise_prompt}

The wealth ranking of the images was conducted using a pairwise comparison strategy, leveraging OpenAI's GPT-4o model to evaluate wealth indicators such as infrastructure quality and visible amenities. 
%The key steps and decisions are outlined below.
%
The total number of images is $N$, one for each DHS survey site. Every pair of images was compared, asking GPT-4o which image represents the wealthier location. The total number of comparisons is ${N \choose 2}$, some of which can end up as \textbf{unresolved}. All comparisons that were not \textbf{unresolved} were used to produce the final ranking. It is worth mentioning that pairwise comparison was chosen not only for methodological reasons, but also because such judgments are cognitively simpler, easier to explain, and easier to audit than absolute scoring approaches.

%Image Loading and Stratified Sampling. The function "load images" loads a CSV file containing image information, including wealth class and image name, enabling stratified sampling of 608 images distributed over 5 classes. Stratified sampling was used to ensure balanced representation of wealth classes, allowing fair comparisons during ranking.

% Pairwise Image Creation. The function "create pairs" generated unique pairs of images for comparison, ensuring exhaustive pairwise comparisons.

%Google Cloud Storage Integration. Signed URLs were generated for each image using Google Cloud Storage, with a caching mechanism implemented to optimize the number of API requests, thus improving performance and minimizing costs.

We used Google Cloud Storage and generated signed URLs for each image, with a caching mechanism implemented to optimize the number of API requests, thus improving performance and minimizing costs. The pairs were compared using OpenAI's GPT-4o model by assessing wealth indicators such as building materials, number of floors, visible greenery, and amenities. The model was asked to judge which of the two locations appeared relatively wealthier based on visible indicators. One retry was done if the model provided an ambiguous response or if the API call failed. Here ``ambiguous'' means that the model responded with either both images or none of the images in the answer. If the response stayed ambiguous or the request failed after the retry, the comparison was considered unresolved. The execution order of the comparisons is set to be random.
%
%Image wealth ranking using OpenAI API. The "rank pair" function compares two images using OpenAI's GPT-4 model by assessing wealth indicators such as building materials, number of floors, visible greenery, and amenities. If the model provided an ambiguous response, a retry mechanism was implemented, and in the case of persistent ambiguity, a deterministic method based on the filename was used to break ties.
%
The prompt used to compare the images is presented in Listing~\ref{listing:prompt}.

\begin{figure*}

\begin{lstlisting}[
basicstyle=\normalfont\ttfamily,
caption={The prompt used for comparing the images.},
captionpos=b,
label={listing:prompt}
]
Compare the following two images based on specific indicators of wealth such as:
1) Quality of infrastructure (e.g., building materials, visible road conditions),
2) Number of floors in the buildings,
3) Amount of visible greenery or well-maintained areas,
4) Presence of visible amenities (e.g., paved roads, power lines).
Please respond with either 'Image 1 is wealthier' or 'Image 2 is wealthier'.
\end{lstlisting}

\end{figure*}

% Threading and Pairwise Ranking: To speed up the process, pairs were processed concurrently using multiple threads (ThreadPoolExecutor), significantly reducing the time required to rank multiple image pairs. Transitivity was applied to propagate pairwise relationships and minimize redundant comparisons.

%\textcolor{blue}{We performed comparisons between each possible pair of images. For every comparison, a prompt was given to ChatGPT, and its answer was analyzed to determine which image represents a wealthier place. If none of the images were mentioned in the answer, it was taken as a tie and was not used in the calculation of ranking.}

%Saving Complete Ranking: After pairwise ranking, the results were saved in a structured CSV file, including rank, file name, and wealth class, to facilitate traceable and interpretable downstream analysis. The final ranking was limited to XXXX images.

%{\bf This section needs rewriting...there are things here that we cannot have done...}

\subsection{Producing the final ranking}

Inferring a total ranking from pairwise comparisons is done by the Bradley-Terry model. For solving this, we used the Iterative Luce Spectral Ranking (I-LSR)~\cite{maystre_fast_2015} algorithm as implemented in the choix Python library (\url{http://choix.lum.li/en/latest/}). The I-LSR algorithm searches for the stationary distribution of a Markov chain with a transition matrix that encodes the pairwise comparisons. We also tried two other algorithms for this in the same library: Luce Spectral Ranking (LSR)~\cite{maystre_fast_2015} and Rank Centrality (RC)~\cite{NegahbanOS2012}.

\subsection{Convolutional Neural Network (CNN)}

For comparing the ranking results to a state-of-the-art image-based method, we report how well a CNN model does on ranking the wealth index for the DHS clusters. The CNN was trained using transfer learning on 10 m/pixel-resolution images from Sentinel-2, with an area of about $2\times2$ square kilometers. The training method and data sets used are described in detail in Sarmadi et al.~\cite{sarmadi_towards_2023}. The training followed very closely the methodology described by Xie et al~\cite{XieEtAl2016, JeanEtAl2016}, but used lower-resolution images. Our experience is that using the lower-resolution Sentinel-2 images yields very similar results to using the higher-resolution Google images.

\subsection{Random Forest regressor}

To compare the ranking with domain expert performance, we use a Random Forest regressor~\cite{breiman_random_2001} using expert-defined features detected in the images. To get a fair evaluation of this model, the training and testing of the Random Forest have been done using k-fold cross-validation. The features and the profile for the domain experts are described in~\cite{sarmadi_human_2024}, where it is also shown that this Random Forest model is better at ranking the images by wealth than the domain experts are themselves when asked to judge the wealth level from the satellite image.

The CNN and random forest regressors are included as reference baselines to contextualize the LLM-based ranking, rather than as competing approaches or primary contributions of the study.

\subsection{Evaluation measures}

Two evaluation measures are reported in the results section. The first is Spearman's rank correlation ($\rho$), the correlation between two lists of ranked values. Spearman's rank correlation varies between -1 and +1, with the latter indicating perfect alignment. A value of 0 indicates no alignment, i.e., random guessing. 
The other is Matthew's correlation coefficient (MCC)~\cite{Matthews75}, or $\phi$ coefficient~\cite{Yule12}, which indicates how well values in a confusion matrix match. In the binary classification case, i.e., when the confusion matrix is $2 \times 2$, MCC varies between -1 and +1, with +1 indicating perfect classification. An MCC value of 0 indicates random guessing. 
We use Gorodkin's generalization of MCC~\cite{Gorodkin04} for the multiclass case with a $5 \times 5$ confusion matrix when wealth indices (true and predicted) are grouped into quintiles. For the multiclass MCC, a value of +1 also indicates a perfect classification and 0 indicates random guessing. However, the minimum is not always -1; it can fall between -1 and 0.

For significance testing, the standard deviations are estimated using the bootstrap method~\cite{Efron79}. In all tests, the significance level is 0.05.

\bibliographystyle{ieeetr}
\bibliography{references}

\section{Acknowledgements}

The authors thank Ibrahim Wahab for his efforts in preprocessing the DHS dataset, which facilitated our analyses. The authors have taken assistance from LLM-based AI tools to brainstorm ideas for this paper's related work, future directions, and limitations, as well as parts of the software implementation and experiment design. All outputs were critically revised by the authors.

\section{Funding}
This work was supported by the Swedish Research Council (grant number 2019-04253) and Riksbankens Jubileumsfond (grant number MXM19-1104:1).

\section{Author contributions statement}

Conceptualization: O.H., T.R.; Software Implementation: H.S., O.H.; Formal analysis and Visualization: H.S., T.R.; Investigation: H.S.; Supervision: O.H., T.R., M.O.; All authors have contributed to developing the methodology and validation, and drafting and revising the manuscript.

\section{Data Availability}
DHS Survey data for Tanzania in 2015 is available at: \url{https://www.dhsprogram.com/data/dataset/Tanzania\_Standard-DHS\_2015.cfm}. The Sentinel-2 daylight images were downloaded from the COPERNICUS/S2 image collection of the Google Earth Engine data catalog: \url{https://developers.google.com/earth-engine/}. The preprocessed Sentinel-2 images used in this study could be downloaded from \url{https://doi.org/10.7910/DVN/PPWAFG}. The code and data used for analysis in this paper are available at the public repository \url{https://github.com/HSarham/PovertyRankingChatGPT}.

\section{Competing interests}
The authors declare no competing interests.

\end{document}